\documentclass[conference]{IEEEtran}
\IEEEoverridecommandlockouts

\usepackage{lipsum}
\usepackage{multirow}
\usepackage{array} 
\usepackage{bbding}
\usepackage{enumitem}
\usepackage{booktabs}
\usepackage{amsmath}
\usepackage{algorithm}
\usepackage{algorithmic}
\usepackage[hyphens]{url}
\usepackage{hyperref}

\usepackage{cite}
\usepackage{amsmath,amssymb,amsfonts}
\usepackage{algorithmic}
\usepackage{graphicx}
\usepackage{textcomp}
\usepackage{xcolor}
\def\BibTeX{{\rm B\kern-.05em{\sc i\kern-.025em b}\kern-.08em
    T\kern-.1667em\lower.7ex\hbox{E}\kern-.125emX}}
\begin{document}

\title{QDA-SQL: Questions Enhanced Dialogue Augmentation for Multi-turn Text-to-SQL\\
}
\author{
\IEEEauthorblockN{1\textsuperscript{st} Yinggang Sun}
\IEEEauthorblockA{\textit{School of Cyberspace Science} \\
\textit{Harbin Institute of Technology}\\
Harbin, China \\
23b903085@stu.hit.edu.cn}
\and
\IEEEauthorblockN{2\textsuperscript{nd} Ziming Guo}
\IEEEauthorblockA{\textit{School of Computer Science and Technology} \\
\textit{Harbin University of Science and Technology}\\
Harbin, China \\
orlosziming@outlook.com}
\and
\IEEEauthorblockN{3\textsuperscript{rd} Haining Yu}
\IEEEauthorblockA{\textit{School of Cyberspace Science} \\
\textit{Harbin Institute of Technology}\\
Harbin, China \\
yuhaining@hit.edu.cn}
\and
\IEEEauthorblockN{4\textsuperscript{th} Chuanyi Liu}
\IEEEauthorblockA{\textit{School of Cyberspace Science} \\
\textit{Harbin Institute of Technology, Shenzhen}\\
Shenzhen, China \\
liuchuanyi@hit.edu.cn}
\and
\IEEEauthorblockN{5\textsuperscript{th} Xiang Li}
\IEEEauthorblockA{\textit{School of Cyberspace Science} \\
\textit{Harbin Institute of Technology}\\
Harbin, China \\
22B903068@stu.hit.edu.cn}
\and
\IEEEauthorblockN{6\textsuperscript{th} Bingxuan Wang}
\IEEEauthorblockA{\textit{School of Cyberspace Science} \\
\textit{Harbin Institute of Technology}\\
Harbin, China \\
1487819688@gg.com}

\and
\IEEEauthorblockN{7\textsuperscript{th} Xiangzhan Yu}
\IEEEauthorblockA{\textit{School of Cyberspace Science} \\
\textit{Harbin Institute of Technology}\\
Harbin, China \\
yxz@hit.edu.cn}
\hspace{9cm}
\and
\IEEEauthorblockN{8\textsuperscript{th} Tiancheng Zhao}
\IEEEauthorblockA{\textit{School of Computer Science and Technology} \\
\textit{Harbin University of Science and Technology}\\
Harbin, China \\
2055525995@qq.com}

}

\maketitle

\begin{abstract}
Fine-tuning large language models (LLMs) for specific domain tasks has achieved great success in Text-to-SQL tasks. However, these fine-tuned models often face challenges with multi-turn Text-to-SQL tasks caused by ambiguous or unanswerable questions. It is desired to enhance LLMs to handle multiple types of questions in multi-turn Text-to-SQL tasks. To address this, we propose a novel data augmentation method, called QDA-SQL, which generates multiple types of multi-turn Q\&A pairs using LLMs. In QDA-SQL, we introduce a method incorporating validation and correction mechanisms to handle complex multi-turn Text-to-SQL tasks. Experimental results demonstrate that QDA-SQL enables fine-tuned models to exhibit higher performance on SQL statement accuracy and enhances their ability to handle complex, unanswerable questions in multi-turn Text-to-SQL tasks. 
\footnote{The generation script and test set of QDA-SQL are released at https://github.com/mcxiaoxiao/QDA-SQL}

\end{abstract}


\section{Introduction}
Text-to-SQL, a notable challenge in natural language processing (NLP), focuses on translating natural language questions into executable SQL queries. It aims to bridge the gap between natural language and SQL for relational databases, thereby building a natural language interface for databases. Recently, due to their powerful understanding and encoding capabilities, LLMs have become a new paradigm for Text-to-SQL. Fine-tuning open-source LLMs has achieved impressive results \cite{sun2023sql,scholak2021picard} in various Text-to-SQL tasks, such as BIRD \cite{li2024can} and CoSQL \cite{yu2019cosql}.

\begin{figure}
  \centering
  \includegraphics[width=0.46\textwidth]{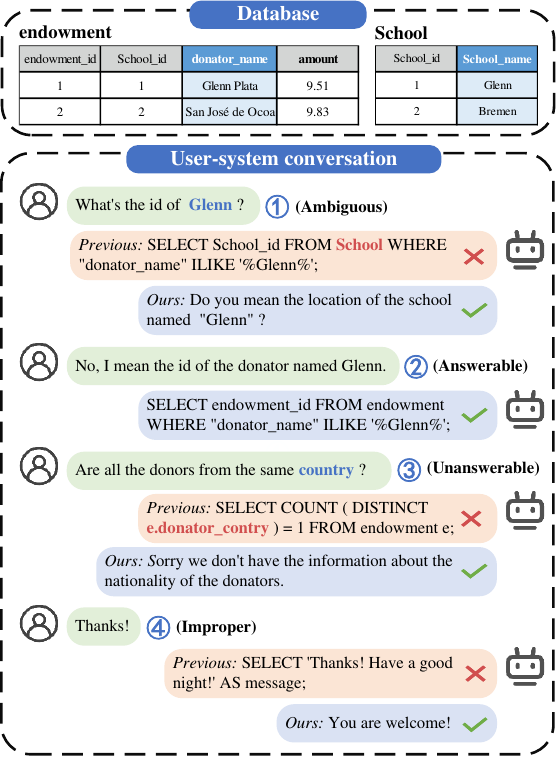}
  \caption{User-LLM dialogues with various question types.}
  \label{fig:User-LLM}
\end{figure}

The dynamism and uncertainty inherent in the real world necessitate effective human-machine interaction. Users often explore data by posing multiple interrelated questions, some of which may not be adequately answered using SQL \cite{wang2023know, 10598154}. Even fine-tuned Text-to-SQL models still exhibit insufficient performance in the above scenarios. As shown in figure \ref{fig:User-LLM}, three types of questions cannot be directly answered by SQL: (1) \textit{Ambiguous}: For instance, the term "Glenn" in the user's first question may map to different columns in different tables, such as "donator\_name" or "school\_name." The system should avoid providing potentially incorrect answers and instead seek clarification from the user.  (2) \textit{Unanswerable}: For instance, if the database lacks information about the "country" of donors in the user's third question, the system needs to articulate the reasons for its inability to answer the question. (3) \textit{Improper}: For instance, the user's fourth question pertains to everyday conversation unrelated to the database, where the system should refrain from responding with SQL.
\renewcommand{\dblfloatpagefraction}{.8}
The above cases reveal the limitations of previous Text-to-SQL models. Existing approaches predominantly focus on enhancing the SQL generation ability of LLMs, without consideration of multiple types of questions. This may result in incorrect responses for questions that cannot be answered using SQL. The weakness may be further exacerbated if LLMs are fine-tuned on datasets that only include Question-SQL samples such as Spider, SParC \cite{yu2019cosql}, and CHASE \cite{guo2021chase}. To enhance the ability of Text-to-SQL systems to handle multiple types of questions, TriageSQL \cite{zhang2020did} and NoisySP \cite{wang2023know} introduce question type recognition and sample generation methods. However, these methods primarily classify question types but lack SQL generation or explanatory features. Additionally, previous data augmentation methods \cite{liu2022augmenting,wang2024conda,zhang2024se}, primarily focus on generating single-turn or multi-turn dialogues that only include SQL queries, without considering multiple question types. Moreover, these methods, which modify existing databases based on preset rules, often yield unnatural questions. Our research endeavors to leverage LLM to refine data augmentation for Text-to-SQL by generating a more diverse and natural set of training samples, thereby enhancing their practical applicability.

In this work, we introduce a new method to automatically generate large batches of Text-to-SQL samples that include multiple types of multi-turn Q\&A, \textbf{Q}uestions Enhanced \textbf{D}ialogue \textbf{A}ugmentation for multi-turn Text-to-SQL named \textbf{QDA-SQL}. With multi-turn Text-to-SQL samples generated by QDA-SQL, fine-tuned LLMs can handle complex multi-turn Text-to-SQL tasks involving various question types. Evaluation across metrics shows our approach's effectiveness.

Our main contributions are summarized as follows:

\begin{itemize}
    \item We propose a new method (i.e., QDA-SQL) to automatically generate large batches of multiple types of multi-turn Text-to-SQL samples, thereby significantly enhancing the robustness of Text-to-SQL LLMs.
    \item We conceptualize the Text-to-SQL inference process as StateFlow and decompose the training data generated by QDA-SQL, improving the model's accuracy and effectiveness in handling complex Text-to-SQL tasks.
    \item We validate the effectiveness of our approach across a range of evaluation metrics, showing that QDA-SQL enhances the model's ability to handle complex multi-turn questions.

\end{itemize}

\section{Related Work}
\subsection{Text-to-SQL}
Text-to-SQL, which transforms natural language queries into SQL queries, is a significant research area with numerous applications. Prominent datasets include single-turn datasets such as Spider BIRD \cite{li2024can}, and multi-turn datasets such as CHASE, SParC, and CoSQL, which serve as benchmarks for evaluating Text-to-SQL systems. Initial studies employed graph neural networks to encode dialogue history and database schema information \cite{hui2021dynamic}, or utilized pre-trained language models (e.g., T5, BERT, BART) to enhance semantic representation. More recent approaches have adopted powerful closed-source LLMs and in-context learning strategies, or have fine-tuned large pre-trained language models to enhance performance.

Recent studies highlight that not all user queries can be effectively addressed using SQL in practical applications \cite{lee2024trustsql, saparina2024ambrosiabenchmarkparsingambiguous, wang2023know}. This is primarily due to the ambiguity in question formulation and the complexity of queries in real-world Text-to-SQL applications \cite{10598154, saparina2024}. To address this issue, datasets such as TriageSQL, NoisySP, and CoSQL have incorporated question type detection, thereby increasing the complexity of the task. Our research aligns with these developments by integrating unanswerable questions into multi-turn Text-to-SQL augmentation, thus creating a more realistic Text-to-SQL task.

\subsection{Data augmentation for Text-to-SQL}
Annotating Text-to-SQL samples is both expensive and time-consuming, as it requires SQL specialists. Consequently, data augmentation has become a widely adopted strategy \cite{zhang2024se, liu2022augmenting}. Early efforts focused on generating single-turn samples using syntax-and-table-aware semantic parsers, copy-based latent variable models \cite{guo2018question}, or predefined transformation rules \cite{wang2024conda, yu2021grappagrammaraugmentedpretrainingtable}. However, these methods often result in unnatural question-answer pairs and fail to address the common multi-turn scenarios encountered in real-world applications.

Previous methods have utilized a self-play strategy for multi-turn generation, where two models simulate the roles of the system and user to generate consecutive question-answer pairs \cite{liu2022augmenting}. To address the generation of multiple question types, some researchers have modified existing database structures to transform answerable samples into ambiguous or unanswerable ones \cite{zhang2020did, wang2023know}. With the advancement of LLMs, the automatic generation of training samples has become increasingly prevalent \cite{ding-etal-2024-data}. Our work was inspired by existing work on data augmentation for semantic parsing and leverage LLMs to generate more natural samples, enabling multi-turn and multi-type interactions, thereby expanding the diversity of the training dataset to develop a more realistic and applicable Text-to-SQL data augmentation strategy.

\begin{table*}
\centering
\caption{Relation categories between the previous question and the current question. The entities (\textbf{bold}), properties(\textit{italics}), and constraints (\underline{underlined}) are highlighted in each question.}
  \begin{tabular}{p{5em}<{\centering} p{15em}<{\centering} p{25em}<{\centering}} 
    \toprule
    \begin{tabular}{p{5em}<{\centering}}\textbf{Relation Category}\end{tabular}    
    &\begin{tabular}{p{15em}<{\centering}}\textbf{Description}\end{tabular}
    &\begin{tabular}{p{25em}<{\centering}}\textbf{Example} \end{tabular}
    \\
    \midrule
    \begin{tabular}{p{5em}<{\centering}}Constraint Refinement\end{tabular}    
    &\begin{tabular}{p{15em}}Ask for the same type of entity as the previous question with a different constraint.\end{tabular}
    &\begin{tabular}{p{25em}}Prev\_Q: Which \textbf{major} has the \underline{fewest}\underline{ students}?\\Cur\_Q: What is the \underline{most popular one}?\end{tabular}
    \\
    \midrule
    \begin{tabular}{p{5em}<{\centering}}Topic Exploration\end{tabular}        
    &\begin{tabular}{p{15em}}Ask for other properties about the same entity as the previous question.\end{tabular}      
    &\begin{tabular}{p{25em}}Prev\_Q: What is the capacity of  \textbf{Anonymous Donor Hall}?\\Cur\_Q: List \textit{ all of the amenities} that \textbf{it} has.\end{tabular}
    \\
    \midrule
    \begin{tabular}{p{5em}<{\centering}}Participant Shift\end{tabular}        
    &\begin{tabular}{p{15em}}Ask for the same property about another entity.\end{tabular}        
    &\begin{tabular}{p{25em}}Prev\_Q: Tell me the \textit{rating} of \textbf{the episode named "Double Down"}.\\Cur\_Q: How about for \textbf{"Keepers"}?\end{tabular}
    \\
    \midrule
    \begin{tabular}{p{5em}<{\centering}}Answer Exploration\end{tabular}        
    &\begin{tabular}{p{15em}}Ask for a subset of entities from a previous question's answer or inquire about a specific entity mentioned in the answer.\end{tabular}
    &\begin{tabular}{p{25em}}Prev\_Q: Please list all the different \textbf{department} \textit{names}. \\Cur\_Q: What is the \textit{average salary} of all instructors in the \textbf{Statistics department}?\end{tabular}
    \\
    \bottomrule
  \end{tabular}
 
  \label{table:Relation}
\end{table*}

\section{Preliminary}
\subsection{Thematic Relation}
Understanding the thematic relations between the current question and previous interactions is crucial for the model to respond in multi-turn dialogues accurately. Drawing from the previous research \cite{bertomeu2006contextual}, which analyzed real interactions involving 30 participants over 2,534 dialogue turns in a Wizard-of-Oz experiment, and incorporating further refinements from SParC \cite{yu2019sparc}, we categorize the relationships between current and prior questions into four distinct types, as shown in Table \ref{table:Relation}. These four categories are designed to encompass various real-world relations observed in natural dialogues, addressing diverse contextual phenomena that occur in practical multi-turn Text-to-SQL interactions. The first three Thematic Relation types pertain to the relationship between a question and a previous question, while the last type pertains to the relationship between a question and the answer to the previous question. By adopting these comprehensive categories in our data augmentation process, we enhance the diversity of the augmented data by considering multiple relationship types.

\begin{table*}
\caption{Question-Answer Types in QDA-SQL. \textit{Note}(italics) explains the rationale for the classification and the requirements for the system's response. \textbf{Bold} font indicates the problematic column description.}
  \centering
  \begin{tabular}{p{4em}<{\centering} p{4em}<{\centering} p{36em}<{\centering}} 
    \hline
    \begin{tabular}{p{4em}<{\centering}}\textbf{Question Category}\end{tabular}    
    &\begin{tabular}{p{4em}<{\centering}}\textbf{Answer Category}\end{tabular}
    &\begin{tabular}{p{30em}<{\centering}}\textbf{Example} \end{tabular}
    \\
    \hline
    \begin{tabular}{p{4em}<{\centering}}Ambiguous\end{tabular}    
    &\begin{tabular}{p{4em}<{\centering}}Clarify\end{tabular}
    &\begin{tabular}{p{36em}}Q: What's the id of \textbf{Glenn}?
    \\A: Do you mean the location of the school named "Glenn"?
    \\ \textit{Note:} The token "Glenn" is ambiguous as it may appear in 
multiple locations within the tables. The system needs to request clarification from the user.\end{tabular}
    \\
    \hline
    \begin{tabular}{p{4em}<{\centering}}Answerable\end{tabular}        
    &\begin{tabular}{p{4em}<{\centering}}SQL\end{tabular}      
    &\begin{tabular}{p{36em}}Q: No, I mean the id of the donator named Glenn.
    \\A: SELECT endowment\_id FROM endowment WHERE "donator\_name" LIKE "Glenn\%";
    \\ \textit{Note:} The question is clear, and the queried column exists. The system needs to provide an SQL query based on the context.\end{tabular}
    \\
    \hline
    \begin{tabular}{p{4em}<{\centering}}Unanswerable\end{tabular}        
    &\begin{tabular}{p{4em}<{\centering}}Sorry\end{tabular}        
    &\begin{tabular}{p{36em}}Q: Are all the donors from the same \textbf{country}?
    \\A: Sorry we don't have the information about the nationality of the donators.
    \\ \textit{Note:}  The question involves nationality information that is missing from the database. The system needs to explain to the user the reason for this absence.\end{tabular}
    \\
    \hline
    \begin{tabular}{p{4em}<{\centering}}Improper\end{tabular}        
    &\begin{tabular}{p{4em}<{\centering}}Regular\end{tabular}
    &\begin{tabular}{p{36em}}Q: Thanks!
    \\A: You are welcome!
    \\ \textit{Note:}  The question does not pertain to querying a database, a regular system response will suffice.\end{tabular}
    \\
    \hline
  \end{tabular}
\label{table:types}
\end{table*}

\subsection{Question-Answer Type}
Question types can be categorized into four types: Answerable, Unanswerable, Ambiguous, and Improper. Among these, Ambiguous and Unanswerable contain questions that present particular challenges. Ambiguous questions include (1) Column Ambiguity, where certain terms in the question can map to multiple columns, and (2) Value Ambiguity, where certain terms in the question can map to multiple cell values in the table. Unanswerable questions include (1) Column Unanswerable, where the concepts mentioned in the question do not exist in the table's columns. (2) Value Unanswerable, where the cell values mentioned in the question do not exist in the table. and (3) Out of Scope, where the question exceeds the system's operational scope, such as attempting to invoke a web search within an SQL system. Table \ref{table:types} contains examples and detailed descriptions of each question-answer type.

\begin{figure*}
  \centering
  \includegraphics[width=0.82\textwidth]{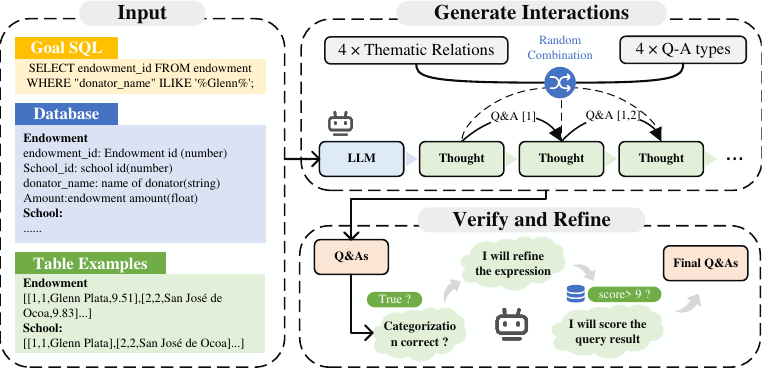}
  \caption{Overview of QDA-SQL processes.}
  \label{fig:experiments2}
\end{figure*}

\label{stateflow}

\begin{figure*}
  \centering
  \includegraphics[width=0.82\textwidth]{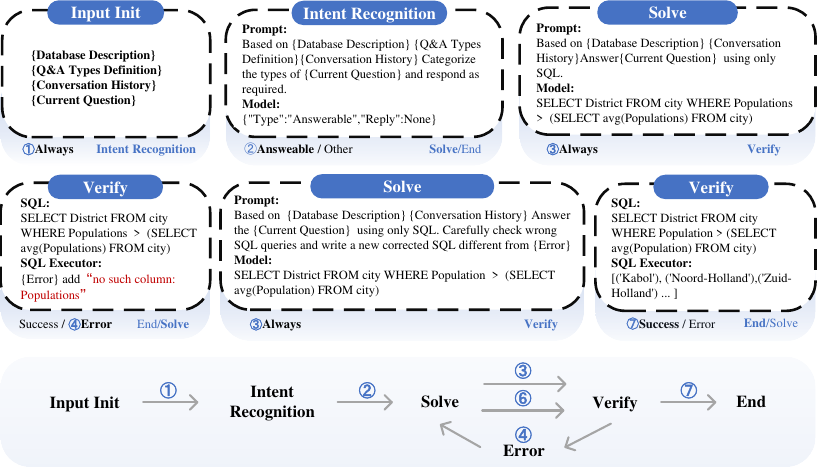}
  \caption{Overview of StateFlow processes}
  \label{fig:experiments3}
\end{figure*}

\section{Methods}
\subsection{Task Formulation}
The problem under consideration is formulated as follows: Given a dialogue context \(C_t\), a current user query \(Q_t\), and a database \(D\) at each turn \(t\) (\(1 \leq t \leq N\)), the Text-to-SQL system is required to identify the type of the user's query \(T_t\) and provide the appropriate response \(R_t\). The response \(R_t\) varies depending on the query type. It may be an SQL query, a clarification question to the user, a natural language response, or an explanation of why the query cannot be answered.
\begin{equation}
  \label{eq:example1}
    \mathrm{p}\left(\mathcal{T}_{\mathrm{t}}, \mathcal{R}_{\mathrm{t}} \mid \mathcal{Q}_{\mathrm{t}}, \mathcal{C}_{\mathrm{t}}, \mathcal{D}\right)
\end{equation}

\subsection{Generate Interactions}
As shown in Figure \ref{fig:experiments2}, we construct a Chain-of-Thought (CoT) process 
 \cite{wei2022chain} and randomly combine a Q-A Type and a thematic relation, thereby guiding the LLM to generate diversified and contextually related multi-turn Q\&A pairs.
\paragraph{Goal SQL} We use the original dataset's SQL as Goal SQL. We enable LLMs to iteratively generate contextually relevant questions and answers. Each interactive exchange is aligned with the Goal SQL. This method enhances the quality and relevance of the generated multi-turn Q\&A pairs, aligning them more closely with real-world application scenarios.
\paragraph{Random Combination} To enhance diversity in multi-turn interactions, we incorporate randomized thematic relation types and Q-A Types during the Q\&A generation process in QDA-SQL. Without specific guidance, the generated Q\&A pairs tend to be limited to common thematic relations and Q-A Types. To counter this, Drawing on the content of previous rounds and the chosen thematic relation, QDA-SQL randomly assigns a predefined thematic relation and Q-A Type to each round. This strategy directs the LLM to produce richer and more varied Q\&A pairs. Furthermore, if a generated Q\&A pair is deemed Improper and suggests an end to the conversation, we cease further generation to maintain the dialogue's natural flow and coherence. This randomization approach and precise control effectively enable the LLM to create high-quality, diverse, and logically coherent multi-turn Q\&A samples.
\paragraph{CoT}  The CoT employs Gemini Pro as a superior LLM for knowledge distillation to guide the generation process. It facilitates \(N\) rounds of diverse Q\&A pairs creation, grounded in schemas, conversation history, and SQL queries, ensuring coherence through contextual reference. To enhance answer quality, five answerable questions accompanied by SQL queries are validated through database execution. Unexecutable SQL queries are filtered out, and the LLM is prompted for complex queries featuring nested structures and multi-table joins, thereby introducing iteration variations to enrich dialogue diversity.

\subsection{Verify and Refine}
The LLM's sample generation based on classification may overlook correctness, leading to misaligned and erroneous results. To improve sample quality, we designed the Verify and Refine process, which directs the LLM through three key steps as shown in Figure \ref{fig:experiments2}.

\begin{enumerate}
\item \textbf{Verifying Alignment}: The LLM checks each Q\&A pair for alignment with its intended question-answer type and removes misaligned pairs to uphold classification standards. For unanswerable samples, we use a stricter setup, which requires LLM with higher temperatures to repeat the check 5 times and pass if they all match.
\item \textbf{Expression Optimization}: Inspired by Self-refine \cite{madaan2024self}, we iteratively refine user queries and system responses using feedback to enhance their readability and naturalness, ensuring they closely resemble natural language.

\item \textbf{Scoring SQL Execution}: For answerable type questions, the LLM scores the SQL execution for the generated answers, ranging from 0 to 10. Higher scores indicate better quality and alignment for the user's question. Answers with scores above a set threshold are retained for their high accuracy and reliability.
\end{enumerate}

Following the above steps, we can obtain validated multi-turn Q\&A samples.

\subsection{StateFlow Design}

\begin{table*}[h]
\caption{Cross-domain multi-turn Text-to-SQL datasets and enhanced data statistics.}
\centering
\begin{tabular}{cccccccc} 
\toprule
\textbf{Dataset} & \textbf{Multi-Type Q} & \textbf{\# Dialogues} & \textbf{\# Turns} & \textbf{Avg. \# Q turns} & \textbf{\# Databases} & \textbf{\# Tables}  \\ \midrule
SParC            & \XSolidBrush                     & 4,298                 & 12,726      & 3.0            & 200       & 1,020                              \\ 
CoSQL            & \Checkmark                     & 3,007                 & 15,598     & 5.2              & 200         & 1,020                             \\ 
CHASE            & \XSolidBrush                     & 5,459                 & 17,940        & 3.3                 & 280       & 1,280                       \\ \midrule
Enhanced Data            & \Checkmark                       & 10,874                 & 65,393            & 6.0        & 200       & 1,020                            \\ \bottomrule
\end{tabular}
  
  \label{table:Comparison}
\end{table*}

We strive to bolster the system's capacity for comprehensive query handling to effectively handle multiple types of questions in text-to-SQL tasks. Building on the work of StateFlow \cite{wu2024stateflow}, we conceptualize the text-to-SQL reasoning process as a state machine model named StateFlow. As shown in Figure \ref{fig:experiments3}, we define the following states and transition rules for implementing the Text-to-SQL system. (1) Initial: Load database info, Q\&A history, question types, user input, and initialize an error log. (2)Intent Recognition: Classifies and parses the user's question.   If the question can be answered via SQL, proceed to the Solve state;   otherwise, respond to the user and then end the interaction. (3) Solve: Generate SQL syntactically compliant. (4) Verify: Checks if the query executes successfully.  If the query executes successfully, it transitions to the End state, completing the task.  If not, it logs errors and redirects to the Solve state to correct the query.

This dynamic prompting and state management mechanism ensures that the LLM receives pertinent guidance at each step, facilitating a structured problem-solving process. The Verify stage detects errors, triggers corrections, and boosts robustness.

To train the LLM for reasoning within the StateFlow model, we organize dialogue samples generated by QDA-SQL to meet the requirements of StateFlow. This method constructs a multi-task training dataset encompassing various tasks in intent recognition and SQL generation. By formatting the dataset to align with the states in the StateFlow model, we ensure logical coherence and effective training.

\section{Data Statistics and Analysis}

As shown in Table \ref{table:Comparison}, we summarize the statistics of several existing multi-turn datasets. We generated 10,874 dialogues, consisting of 65,393 turns, through the QDA-SQL framework, based on the SParC and CoSQL training sets.

\begin{figure}[h]
    \centering
    \includegraphics[width=0.4\textwidth]{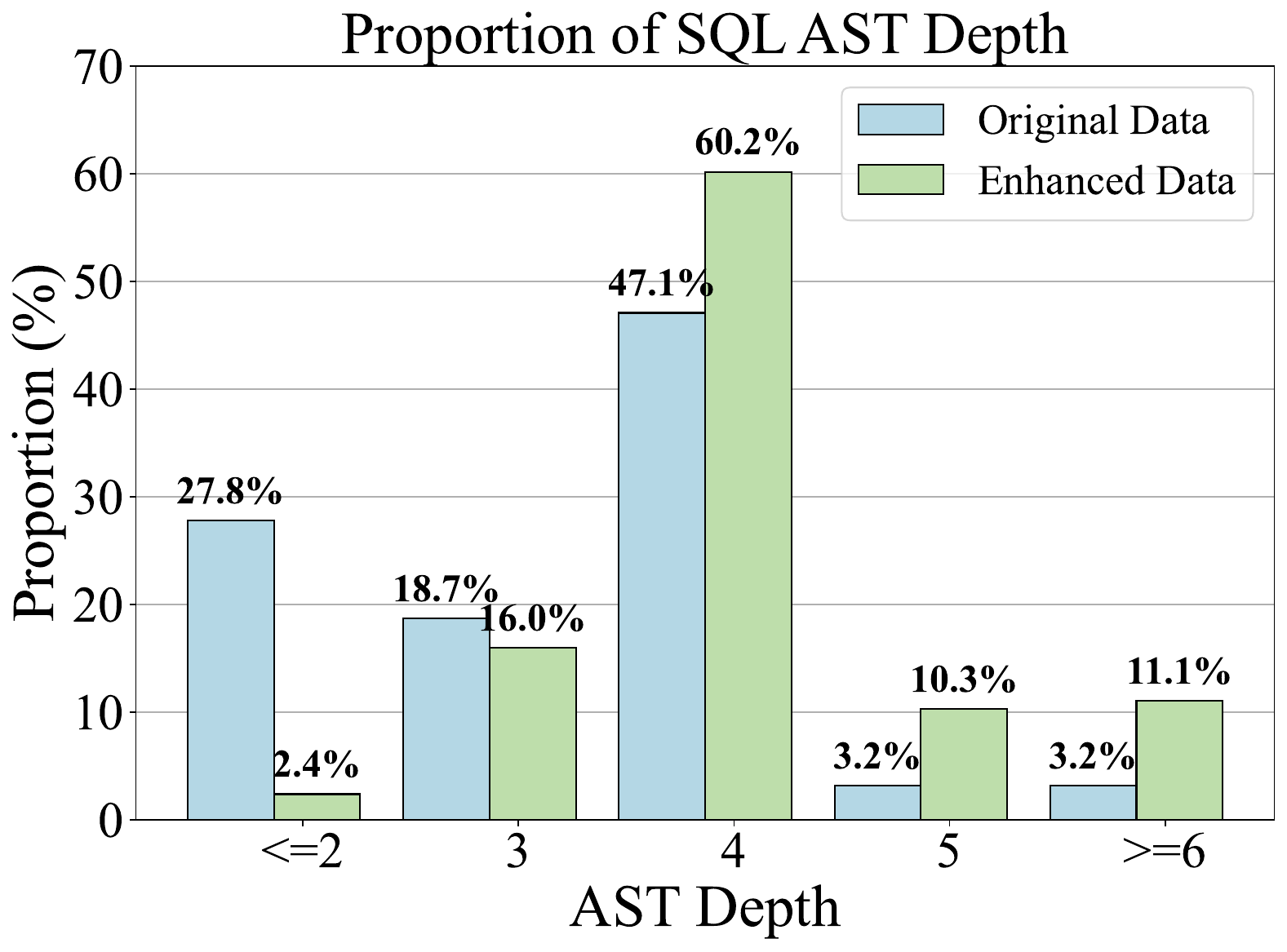}
    \caption{SQL AST depths in enhanced vs. original samples}
    \label{figure:AST}
\end{figure}

\begin{figure}[h]
    \centering
    \includegraphics[width=0.4\textwidth]{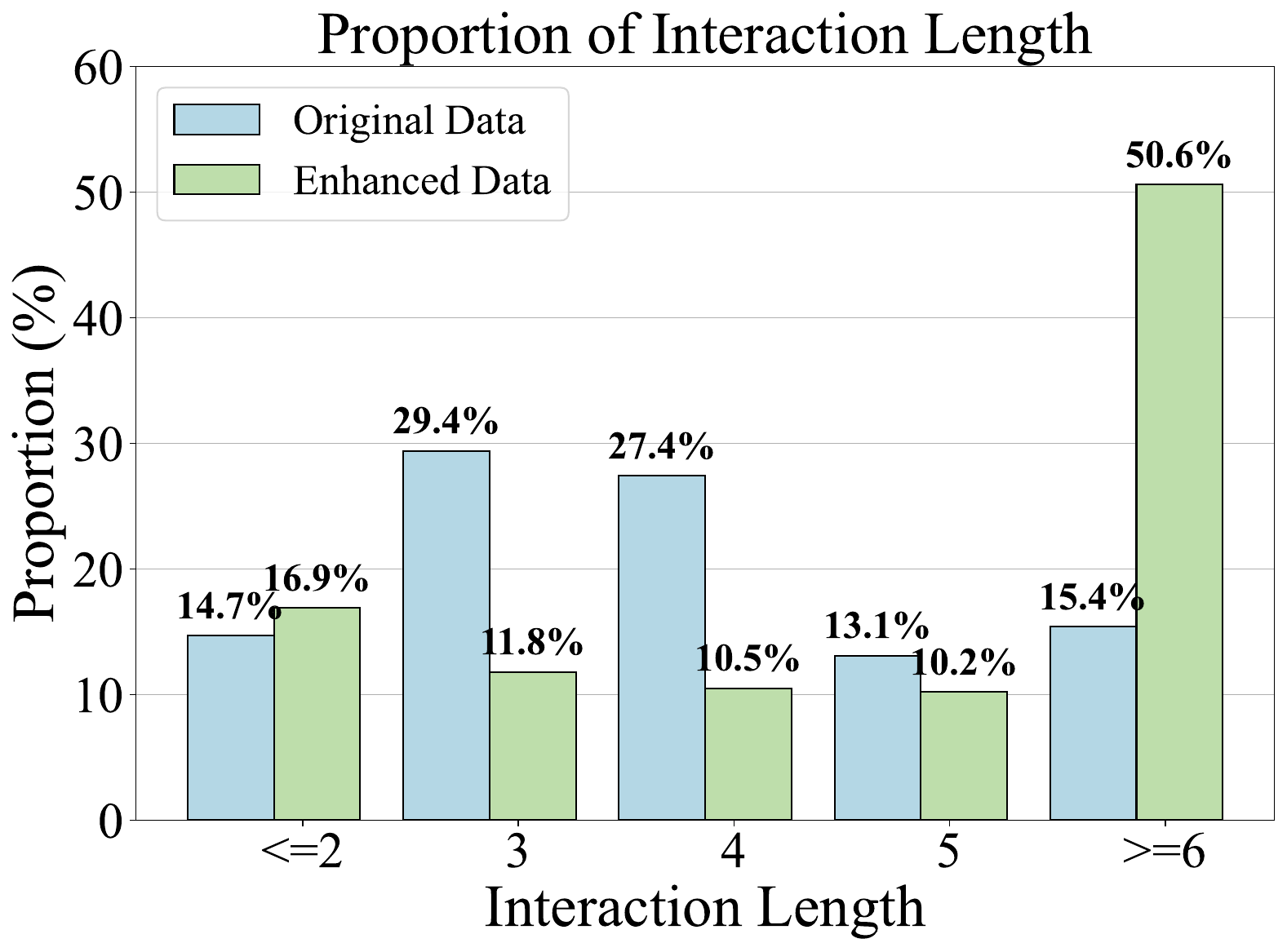}
    \caption{Dialogue lengths in enhanced vs. original samples}
    \label{figure:length}
\end{figure}

Analyzing the SQL output of QDA-SQL involved comparing it to the original dataset based on AST (Abstract Syntax Tree) depth, dialogue length, and augmentation impacts. Figure \ref{figure:AST} illustrates the SQL AST tree depths distribution for enhanced and original samples. The SQL generated in the enhanced dataset demonstrates greater tree depths, suggesting that QDA-SQL can produce more complex queries. This complexity aims to improve the model's ability to handle intricate problems. Figure \ref{figure:length} displays the distribution of dialogue lengths for both enhanced and original samples. These findings suggest that QDA-SQL generates longer dialogues, creating more challenging scenarios. This is intended to improve the model's performance in multi-turn dialogues.

\begin{figure}[h]
    \centering
    \includegraphics[width=0.95\linewidth]{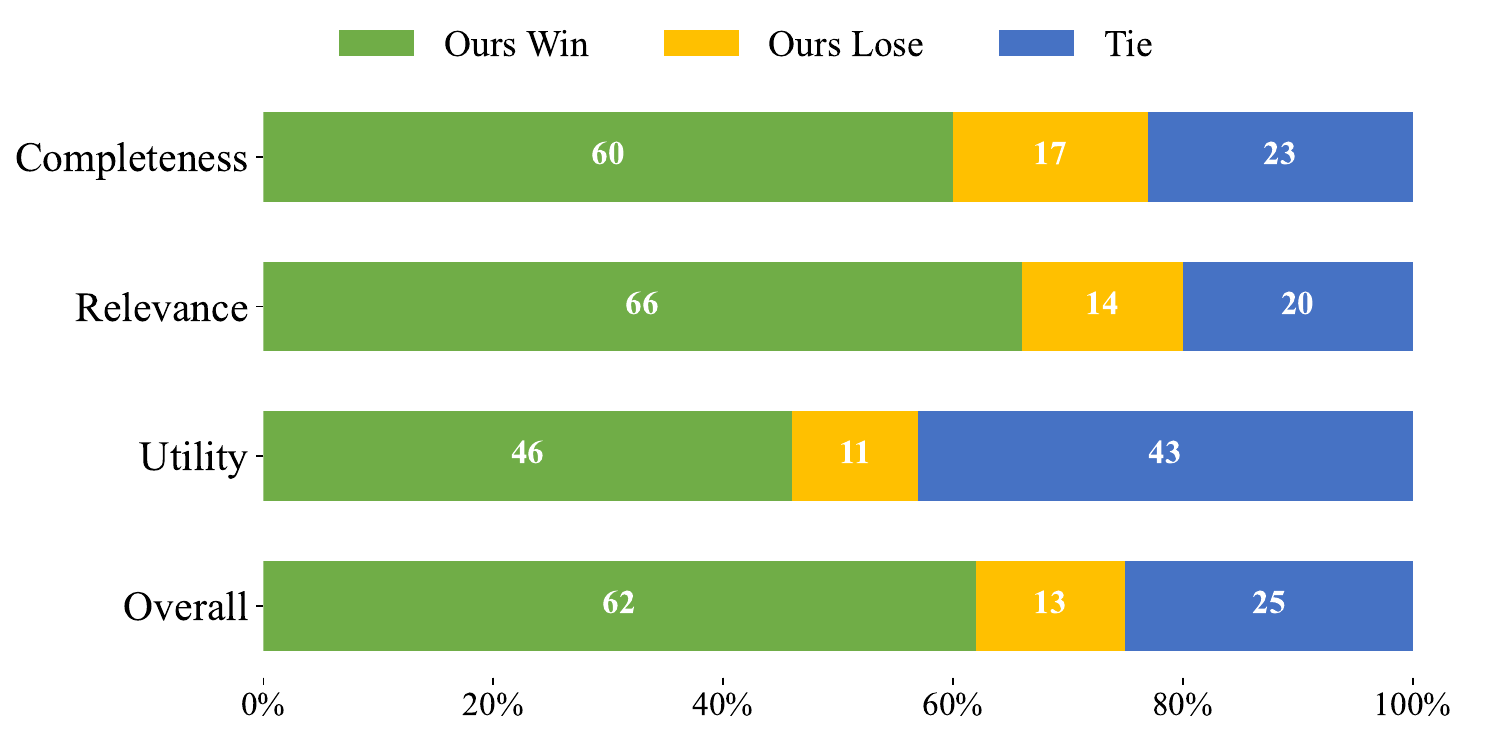}
    \caption{Pairwise comparison of original and QDA-SQL enhanced dataset annotation quality across different criteria.}
    \label{fig:data_comparison_results}
\end{figure}

To assess the further generation quality implemented in QDA-SQL, we conducted a manual review of the automatically filtered results, focusing on both classification accuracy and the quality of the annotated natural language question answering. We compared the samples before and after the Verify and Refine process and found that Gemini Pro accurately identified 94\% (47 out of 50) of the misclassified samples. This indicates that the automatic filtering process is effective in improving the quality of the data, thereby supporting high classification accuracy. Furthermore, we adopted an automatic evaluation framework based on GPT-4, as proposed in previous work \cite{NEURIPS2023_91f18a12, xu2023wizardlmempoweringlargelanguage}, to measure critical aspects such as Completeness, Relevance, and Utility in the annotated natural language question answering. To mitigate order bias \cite{iourovitski2024gradescorequantifyingllm}, we alternated the placement of the original dataset and our enhanced dataset in pairwise comparisons, positioning our dataset first for odd IDs and second for even IDs. As illustrated in Figure \ref{fig:data_comparison_results}, our enhanced dataset shows superior performance, with an overall assessment indicating that 62\% of the QDA-SQL enhanced dataset is regarded as superior to the manually annotated original SParC and CoSQL training sets.

\section{Experiment}
\subsection{Experiment Setup}
We evaluated our methods using two large-scale benchmark datasets: SParC and CoSQL. As the process outlined in section \ref{stateflow}, we used this framework to organize and format the training data, ensuring a balanced distribution of samples between the question type recognition and SQL task. This involved implementing a negative sampling strategy. Additionally, we curated a test set from CoSQL, which we refer to as the Multi-type Multi-turn text-to-SQL test set (MMSQL). For the evaluation, we developed the Accuracy with SQL Matching (AccS) metric. The LLMs are fine-tuned using QLoRA \cite{dettmers2024qlora}; training settings include lora\_rank set to 64, lora\_alpha set to 16, and learning rate set to 5e-5. During inference, the temperature is set to 0.1. Our experiments are conducted on a server with an Intel(R) Xeon(R) Gold 6133 CPU @ 2.50GHz and an NVIDIA A800 80GB PCIe GPU.

\subsection{Evaluation Metrics}
To evaluate our methods, we employ two official metrics: Question Match (QM) and Interaction Match (IM) \cite{yu2019sparc}. QM measures whether the predicted SQL query matches the ground truth at the single-turn level, while IM assesses whether all predicted SQL queries in a multi-turn interaction achieve QM. Furthermore, we present detailed analyses of precision, recall, and F1 scores for user dialogue act prediction across multiple question types. For a comprehensive assessment, we have devised an integrated metric, AccS, that combines question type recognition with SQL query generation metrics. AccS integrates User Dialogue Act Prediction Accuracy (Acc) \cite{yu2019cosql} with Question Match (QM), providing a comprehensive measure of performance.

\paragraph{User Dialogue Act Prediction Accuracy (Acc)} Acc is used to evaluate the system's ability to classify questions. It is defined as the proportion of instances where the predicted type matches the expected reference type among all predicted questions. Formally, for a dataset comprising \( N \) questions, where \( C_i \) denotes the expected classification and \( \hat{C}_i \) represents the predicted classification for the \( i \)-th question, Acc is computed as follows:

\begin{equation}
  \label{eq:example2}
    \text{Acc}=\frac{1}{N} \sum_{i=1}^N \mathrm{I}\left(C_i=\widehat{C}_i\right)
\end{equation}
\\
\noindent 
where \( \mathrm{I} = 1 \) if \( \hat{C}_i \) matches \( C_i \), and \( \mathrm{I} = 0 \) otherwise.
\paragraph{Acc with SQL Matching (AccS)} In the context of a database querying dialogue system, it is imperative not only to accurately classify the type of user query but also to furnish the appropriate SQL queries for those answerable questions. To comprehensively evaluate the system's performance, we enhance Acc by integrating Question Match (QM). This integration results in a comprehensive metric, AccS, which assesses both the system’s proficiency in question classification and its precision in SQL query generation within multi-turn dialogues. For a given set of \( N \) questions, where \( S_i \) denotes the ground truth SQL query and \( \hat{S}_i \) represents the predicted SQL query, AccS is computed as follows:
\begin{equation}
  \label{eq:example3}
    \text{AccS} = \frac{1}{N} \sum_{i=1}^{N} \left\{
    \begin{array}{ll}
    
    \mathrm{I}(C_i = \widehat{C}_i) \cdot \mathrm{QM}(S_i, \widehat{S}_i) & \text{(a)} \\
    \mathrm{I}(C_i = \widehat{C}_i) & \text{(b)}\\
    \end{array}
    \right.
\end{equation}

\begin{equation*}
\begin{array}{ll}

  \text{(a)} &  C_i = \mathrm{'Answerable'} \\

  \text{(b)} & \mathrm{otherwise}

\end{array}
\end{equation*}
\\
\noindent
where \( \mathrm{QM} = 1 \) if \( \hat{S}_i \) matches \( S_i \), and \( \mathrm{QM} = 0 \) otherwise.

AccS can be computed at both the single-turn level and interaction levels. For Interaction Level AccS (IAccS), if all the predicted SQL queries and question types in an interaction are correct, the IAccS score is 1; otherwise, the score is 0.

\subsection{Experiment Result}
In the QDA-SQL process, we generated a training set comprising 10,874 dialogues with a total of 65,393 turns, based on the SParC and CoSQL datasets. We conducted a zero-shot evaluation of several recent state-of-the-art general and domain-specific LLMs using QM and IM metrics with the development sets of SParC and CoSQL.

As shown in Table \ref{table:mainresult}, open-source models without fine-tuning exhibit significantly inferior performance compared to state-of-the-art closed-source models, such as GPT-3.5 Turbo, GPT-4 Turbo, and Gemini-1.5-pro. However, after fine-tuning with the training set we generated, the performance of models such as Nsql-7b \cite{numbersstation2023NSText2SQL}, Sqlcoder-7b-2 \cite{defog2023sql}, Codellama-7b, and Codegemma-7b exceeded that of these closed-source LLMs in both QM and IM. Additionally, to conduct a more detailed evaluation of the real impact of the augmented dataset compared to the original dataset on model performance, we partitioned the QDA-SQL generated dataset and performed an ablation study on representative open-source models (see Section \ref{sec:Ablation}).

\begin{table}
\caption{Model QM and IM performances on the SParC and CoSQL.}

\centering
\begin{tabular}{l|cc|cc}
\toprule

\multirow{2}{*}{\textbf{Model}} & \multicolumn{2}{c|}{\textbf{Sparc}} & \multicolumn{2}{c}{\textbf{CoSQL}} \\
                                & \textbf{QM} & \textbf{IM} & \textbf{QM} & \textbf{IM} \\ \midrule
GPT-3.5 Turbo                               & 34.1         & 17.1        & 32.2        & 10.4        \\
GPT-4 Turbo                                 & 31.4         & 18.8        & 33.6        & 10.8        \\
Gemini-1.5 pro                            & 36.6         & 18.3        & 36.7        & 9.9         \\ \midrule
Nsql-7b                                     & 5.6          & 0.2         & 3.4         & 0.0         \\
+   QDA-SQL                                 & 51.4         & 32.7        & 46.9        & 19.8        \\ \midrule
Sqlcoder-7b-2                               & 0.5          & 0.0         & 0.7         & 0.0         \\
+   QDA-SQL                                 & 53.9         & 35.8        & 48.1        & 22.2        \\ \midrule
Codellama-7b                                & 20.6         & 2.1         & 6.4         & 1.4         \\
+   QDA-SQL                                 & 56.2         & 38.2        & 50.9        & 24.2        \\ \midrule
Codegemma-7b                                & 29.7         & 12.6        & 24.5        & 6.8         \\
+   QDA-SQL                                 & \textbf{57.6}         & \textbf{42.4}        & \textbf{56.5}        & \textbf{28.7}        \\ \bottomrule
\end{tabular}
  
  \label{table:mainresult}
\end{table}

\begin{figure}
    \centering
    \includegraphics[width=0.36\textwidth]{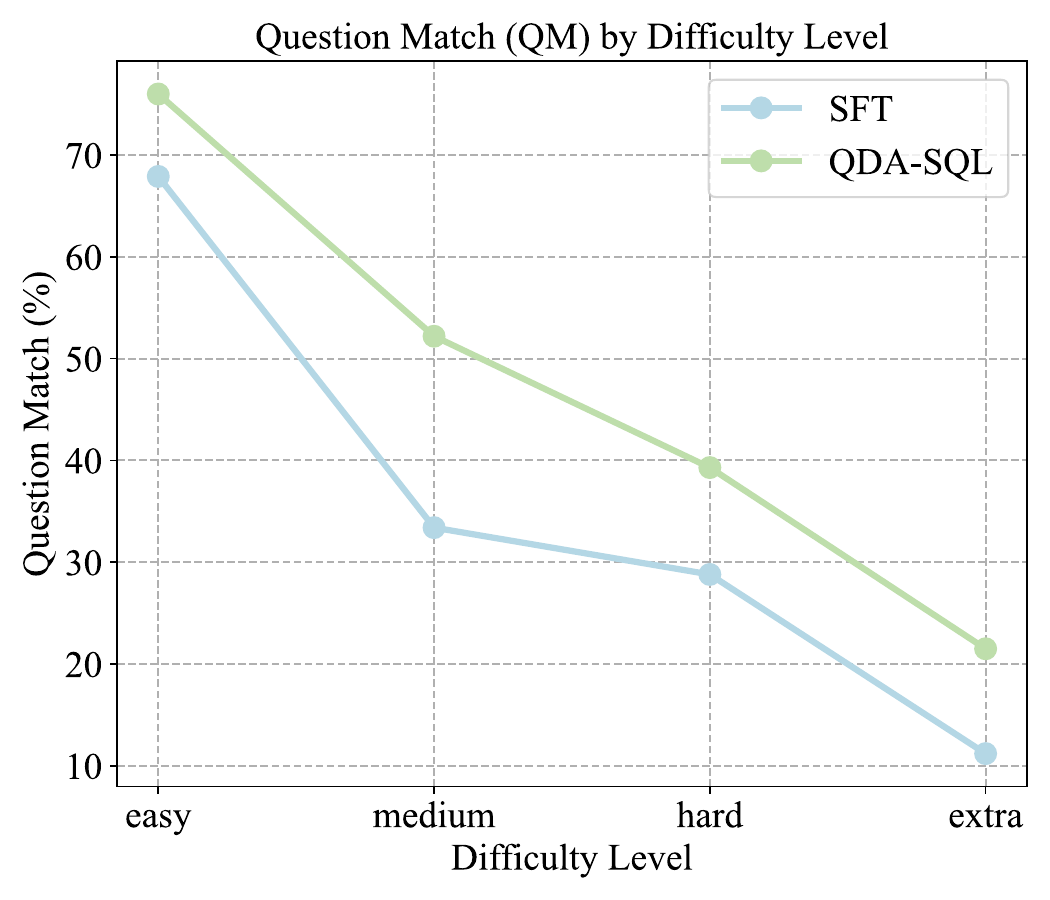}
    \caption{QM performance on SQL of different difficulty levels}
    \label{figure:QMBYDI}
\end{figure}

\begin{figure}
    \centering
    \includegraphics[width=0.36\textwidth]{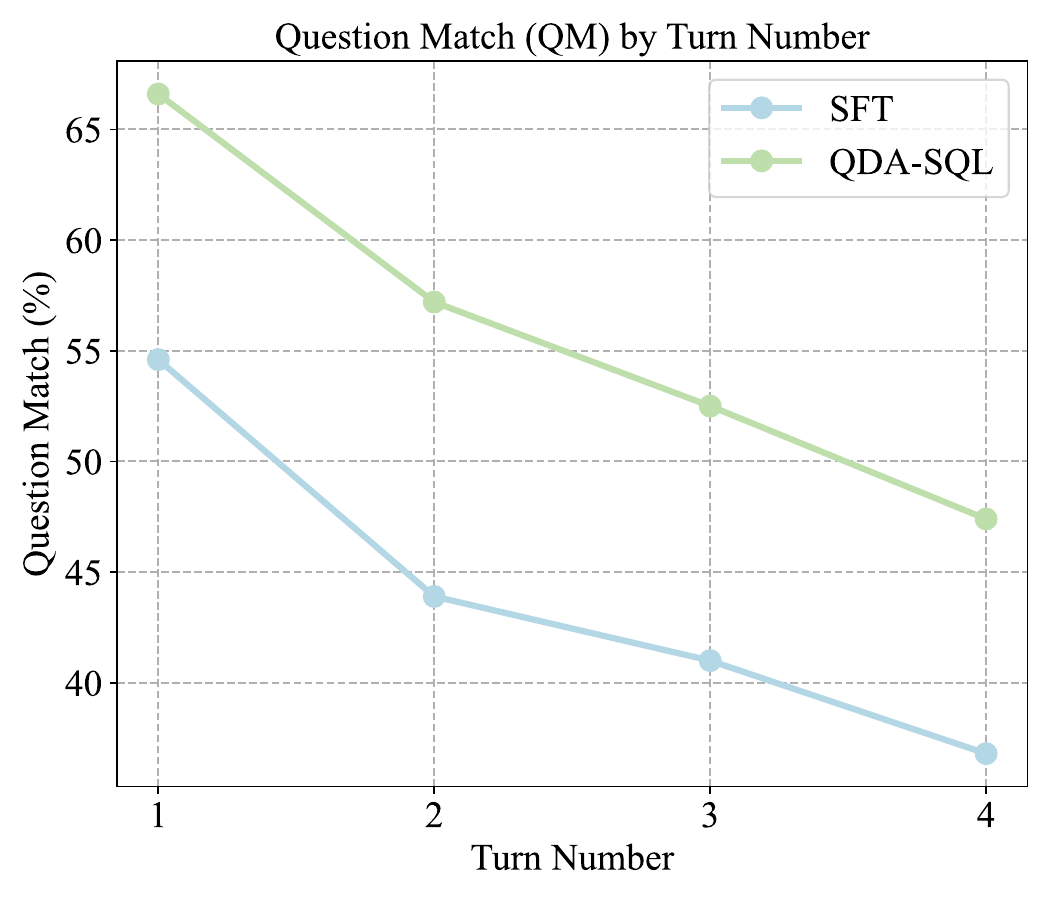}
    \caption{QM performance on problems with different turn numbers}
    \label{figure:QMBYTU}
\end{figure}

To evaluate the impact of samples generated by QDA-SQL on model fine-tuning, we analyzed LLMs trained with augmented data versus those trained with original datasets across various dialogue turns and difficulty levels. These difficulty levels were defined according to the Spider standard, encompassing easy, medium, hard, and extra hard. Figure \ref{figure:QMBYDI} illustrates the QM of SQL outputs for Answerable questions across these difficulty levels, comparing results from models trained with original data to those trained with the enhanced dataset generated by QDA-SQL. Notably, models fine-tuned with the augmented dataset demonstrated significant improvements in QM, especially for questions with higher difficulty. Figure \ref{figure:QMBYTU} presents the QM performance across different turn numbers, contrasting models trained with the original data versus the enhanced dataset. The enhanced training dataset demonstrated marked improvements in QM for multi-turn dialogues across varied difficulty levels, thereby highlighting its efficacy in complex dialogue scenarios.

\subsection{Ablation Study}
\label{sec:Ablation}
In this study, we aimed to conduct a more detailed evaluation of the impact of the augmented dataset as well as the influence of various states within the StateFlow on model performance.

We partitioned the QDA-SQL-generated dataset and performed ablation studies on representative open-source models. According to the results in Table \ref{table:2long} and Table \ref{table:llamaoverall}, the two enhanced partition versions of the QDA-SQL dataset, one containing a complete set of samples with intent recognition and another including only answerable samples, demonstrated superior performance compared to models trained solely on the original dataset. These results confirm the effectiveness of the QDA-SQL as a method for improving model performance on both SQL generation and intent recognition tasks.

As shown in Table \ref{table:llamaoverall} and Table \ref{table:3long}, models trained with QDA-SQL equipped with Intent Recognition samples (\emph{With Intent}) significantly outperform the SFT model, achieving an average F1 score of 57.8. Notably, these \emph{With Intent} models not only exhibit higher intent recognition capabilities but also achieve QM scores comparable to those of the \emph{SQL-only} models. Conversely, models trained with the dataset without Intent Recognition (\emph{SQL-only}) show improvement in QM but face challenges in recognizing diverse question types. Incidentally, datasets generated by QDA-SQL can be utilized to train models focused solely on answerable tasks, thereby enhancing their SQL capabilities. However, as evidenced in the "QDA-SQL (\emph{SQL-only})" section of Table \ref{table:3long}, these models, while proficient in answerable tasks, exhibit significant limitations when addressing questions that cannot be answered using SQL, struggling to effectively recognize unanswerable and ambiguous questions. These findings underscore the effectiveness of multi-task learning strategies in enhancing the model's robustness in complex interactive environments. By excelling in both intent recognition and maintaining strong QM performance, the \emph{With Intent} models demonstrate a balanced proficiency essential for handling a wide range of queries, including those that are unanswerable or ambiguous. For production environments that prioritize SQL scenarios, training LLMs exclusively on SQL data may yield better performance in SQL tasks than incorporating intent classification tasks. However, models trained on datasets that include both intent recognition and SQL generation tasks demonstrate a more balanced performance, effectively handling a wider range of queries.

\begin{table}
\caption{Performance of models trained on different datasets. "SFT" refers to fine-tuning models on the original CoSQL and SparC training set. "With Intent" includes both SQL prediction and intent recognition samples generated by QDA-SQL, while "SQL-only" only include answerable Q\&A samples generated by QDA-SQL.}
\centering

\begin{tabular}{l|cc|cc}

\toprule
\multicolumn{1}{c|}{\multirow{2}{*}{\textbf{Model}}} & \multicolumn{2}{c|}{\textbf{Sparc}} & \multicolumn{2}{c}{\textbf{CoSQL}} \\ 
\multicolumn{1}{c|}{}                       & QM           & IM          & QM          & IM          \\ \midrule
\textbf{Codellama-7b}                               & 20.6         & 2.1         & 6.4         & 1.4         \\
SFT                               & 55.3         & 39.6        & 47.1        & 20.1 \\
QDA-SQL (\emph{With Intent})                                & 56.2         & 38.2        & 50.9        & 24.2 \\
QDA-SQL (\emph{SQL-only})                                 & \textbf{62.5}         & 41.1        & 56.7        & 28.0 \\ \midrule
\textbf{Codegemma-7b}                           & 29.7         & 12.6        & 24.5        & 6.8         \\
SFT                               & 57.5         & 41.2        & 55.1        & 27.3 \\
QDA-SQL (\emph{With Intent})                                & 57.6         & 42.4        & 56.5        & 28.7 \\
QDA-SQL (\emph{SQL-only})                                & 61.3         & \textbf{44.1}        & \textbf{57.3}        & \textbf{30.0}        \\ \bottomrule
\end{tabular}
  
    \label{table:2long}
\end{table}

\begin{table}
\caption{Experimental results on the CoSQL dev set.}
\centering
\begin{tabular}{lm{0.8cm}<{\centering}m{0.6cm}<{\centering}m{0.6cm}<{\centering}m{0.6cm}<{\centering}}
\toprule
\multicolumn{1}{c}{\textbf{Method}} & \textbf{QM}            & \textbf{Avg. Pre} & \textbf{Avg. Rec}   & \textbf{Avg. F1}       \\ \midrule
\textbf{Codellama-7b}                & 6.4           & 51.1           & 46.1          & 46.4          \\
SFT                         & 47.1          & 47.2           & 44.9          & 45.1          \\
QDA-SQL (\emph{With Intent})             & 55.9          & \textbf{60.0}  & \textbf{57.1} & \textbf{57.8} \\
QDA-SQL (\emph{SQL-only})             & \textbf{56.7} & 47.9           & 46.2          & 46.8     \\
\bottomrule
\end{tabular}
 
  \label{table:llamaoverall}
\end{table}

\begin{table}
\caption{Performance of models in recognizing question types by Codellama-7b on the CoSQL dev set. The best-performing average value is highlighted in \textbf{bold}.}
\centering
\begin{tabular}{lcccc}
\toprule
\multicolumn{1}{c}{\textbf{Model}} & \textbf{Precision} & \multicolumn{2}{c}{\textbf{Recall}}                                                                                        & \multicolumn{1}{c}{\textbf{F1}} \\ \midrule
\textbf{SFT}                &           & \multicolumn{2}{c}{}                                                                                              &                        \\
Answerable                  & 82.4      & \multicolumn{2}{c}{99.3}                                                                                          & 90.1                   \\
Unanswerable                & 0         & \multicolumn{2}{c}{0}                                                                                             & 0                      \\
Ambiguous                   & 6.5       & \multicolumn{2}{c}{0.9}                                                                                           & 1.6                    \\
Improper                    & 100.0     & \multicolumn{2}{c}{79.5}                                                                                          & 88.6                   \\
average                     & 47.2      & \multicolumn{2}{c}{44.9}                                                                                          & 45.1                   \\ \midrule
\textbf{QDA-SQL (\emph{With Intent})}      &           & \multicolumn{2}{c}{}          \\
Answerable                  & 90.5      & 94.8                                                       & \multicolumn{2}{c}{92.6}                                                      \\
Unanswerable                & 21.1      & 19.0                                                       & \multicolumn{2}{c}{20.0}                                                      \\
Ambiguous                   & 30.8      & 14.8                                                       & \multicolumn{2}{c}{20.0}                                                      \\
Improper                    & 97.5      & 99.7                                                       & \multicolumn{2}{c}{98.6}                                                      \\
average                     & \textbf{60.0}      & \textbf{57.1}                                                       & \multicolumn{2}{c}{\textbf{57.8}}                                                      \\ \midrule
\textbf{QDA-SQL (\emph{SQL-only}) }     &           & \multicolumn{2}{c}{}                                                                                              &                        \\
Answerable                  & 88.5      & \multicolumn{2}{c}{98.0}                                                                                          & 93.0                   \\
Unanswerable                & 0         & \multicolumn{2}{c}{0}                                                                                             & 0                      \\
Ambiguous                   & 3.2       & \multicolumn{2}{c}{0.9}                                                                                           & 2.4                    \\
Improper                    & 100.0     & \multicolumn{2}{c}{85.1}                                                                                          & 92.0                   \\
average                     & 47.9      & \multicolumn{2}{c}{46.2}                                                                                          & 46.8                   \\  \bottomrule
\end{tabular}
  
  \label{table:3long}
\end{table}

To assess the impact of various states on model performance, we conducted ablation studies. (1) Removing the Verify State: The SQL generated by the model is accepted as the final answer without any verification. (2) Removing Intent Recognition: Without this feature, the model skips the decision-making process for the next step (End or Solve) based on question classification and proceeds directly to Solve, thereby directly addressing the question.

As shown in Table \ref{table:StateFlow_ablation}, removing either state results in a decline in performance. The removal of the Verify state results in lower AccS and IAccS values, and higher error rates, underscoring its critical importance. The removal of the Intent Recognition state leads to reduced IAccS, as the LLM faces difficulties with questions that cannot be answered using SQL. This is consistent with the idea that the Intent Recognition state improves the LLM's capability to comprehend the types of questions posed.

\begin{table}
\caption{Ablation study results for StateFlow states on the MMSQL test set. "No\_Verify" excludes error correction, and "No\_Intent" excludes intent recognition.}
\centering
\begin{tabular}{cccc}
\toprule
\multicolumn{1}{c}{\textbf{Method}} & \textbf{AccS↑} & \textbf{IAccS↑} & \textbf{ERROR \%↓} \\ \midrule
\textbf{QDA-SQL}    & 50.4           & 12.0            & 1.7                \\
No\_Verify                           & 49.9           & 10.6            & 2.5                \\
No\_Intent                           & 29.6           & 0.4             & 2.3                \\ \bottomrule
\end{tabular}

  \label{table:StateFlow_ablation}
\end{table}

\subsection{Error Analysis}
In our study, we randomly selected 50 records from each of the three models: GPT-4, CodeLlama-7B, and CodeLlama fine-tuned with the dataset generated by QDA-SQL for analysis, totaling 150 records. We categorized error reasons into the following classes:
\begin{itemize}
\item \textbf{Semantic understanding errors}: These include insufficient contextual understanding, misinterpretation of the question, and word confusion.

\item \textbf{SQL execution errors}: These arise from issues such as incorrect function usage and data mismatches, leading to execution failures.

\item \textbf{Logical errors}: These involve incorrect relational expressions, conditional judgments, and aggregate operations errors.

\item \textbf{Database comprehension errors}: These include unfamiliarity with the database structure, limitations related to database functions, and incorrect selection of relevant tables.
\end{itemize}

\begin{figure}
\centering
\includegraphics[width=0.45\textwidth]{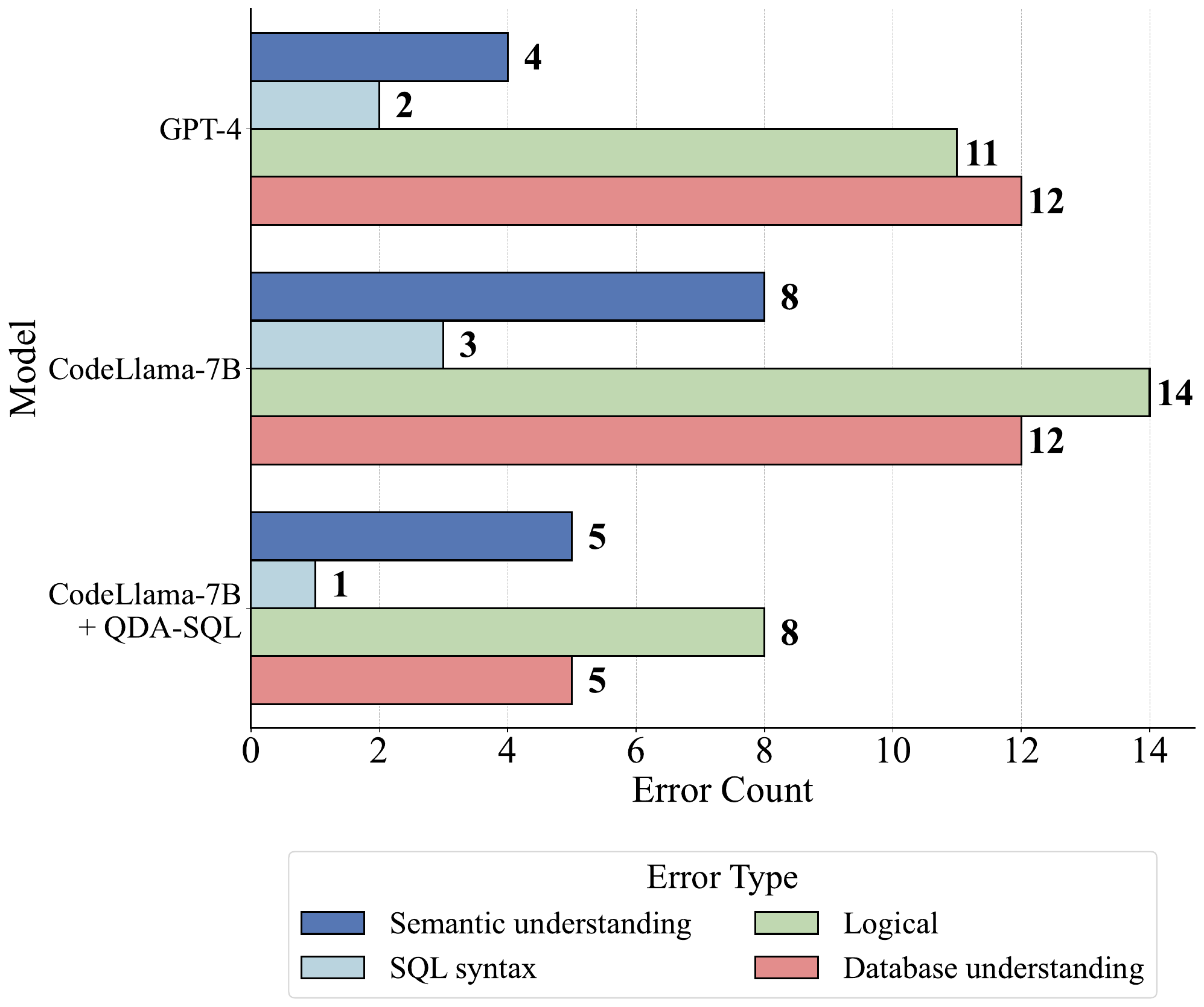}
\caption{Error type distribution across different models.}
\label{fig:error_analysis}
\end{figure}

As illustrated in Figure \ref{fig:error_analysis}, our analysis revealed that both GPT-4 and the unfine-tuned CodeLlama exhibited a significant number of logical and database comprehension errors. This indicates challenges in understanding and manipulating database query logic and schema information. However, GPT-4 showed superior semantic understanding, likely due to its advanced language processing capabilities, which enhance contextual parsing and query intent comprehension. In contrast, the fine-tuned CodeLlama demonstrated the lowest error rate among the models, even with an equivalent number of samples. Notably, the fine-tuned CodeLlama showed a marked reduction in logical and database comprehension errors, highlighting the critical role of fine-tuning in improving model performance on specialized tasks. In conclusion, fine-tuning significantly enhanced the model's domain-specific capabilities, particularly in database comprehension. While closed-source models like GPT-4 demonstrate robust natural language understanding straight out of the box, there is still potential for enhancement in specific domain applications.

\section{Limitations and Future Work}
Our methodology utilizes LLMs to generate samples from Goal SQL, however, creating Goal SQL still necessitates labor-intensive manual annotation. Furthermore, the dialogue samples produced may not cover all conceivable real-world scenarios. Future research will concentrate on automating the generation of diverse Text-to-SQL Q\&A pairs. This encompasses scenarios that require external knowledge and complex reasoning processes. Our objective is to refine augmentation techniques to enhance their applicability, aiming to generate more challenging and varied samples that accommodate a broader range of database structures and querying requirements.

\section{Conclusion}

In this paper, we introduce QDA-SQL, an innovative LLM-based data augmentation method tailored for multi-turn Text-to-SQL tasks. QDA-SQL significantly enriches the diversity and quality of training datasets. By integrating with the StateFlow framework, we have refined the training and inference processes for LLMs, effectively guiding them through multi-step reasoning. Our experimental results indicate a marked improvement in SQL accuracy and more adept handling of a variety of question types, thereby bolstering the accuracy and robustness of LLMs in real-world applications. Moreover, the data augmentation and inference framework we have developed is scalable and can accommodate a wider array of scenarios encountered in real-world Text-to-SQL applications.

\bibliographystyle{IEEEtran}
\bibliography{custom}

@inproceedings{yu2019cosql,
  title={CoSQL: A Conversational Text-to-SQL Challenge Towards Cross-Domain Natural Language Interfaces to Databases},
  author={Tao Yu and Rui Zhang and He Yang Er and Suyi Li and Eric Xue and Bo Pang and Xi Victoria Lin and Yi Chern Tan and Tianze Shi and Zihan Li and Youxuan Jiang and Michihiro Yasunaga and Sungrok Shim and Tao Chen and Alexander R. Fabbri and Zifan Li and Luyao Chen and Yuwen Zhang and Shreya Dixit and Vincent Zhang and Caiming Xiong and Richard Socher and Walter S. Lasecki and Dragomir R. Radev},
  booktitle={Conference on Empirical Methods in Natural Language Processing},
  year={2019},
  url={https://api.semanticscholar.org/CorpusID:202565697}
}

@misc{sun2023sql,
      title={SQL-PaLM: Improved Large Language Model Adaptation for Text-to-SQL (extended)}, 
      author={Ruoxi Sun and Sercan {\"O}. Arik and Alex Muzio and Lesly Miculicich and Satya Gundabathula and Pengcheng Yin and Hanjun Dai and Hootan Nakhost and Rajarishi Sinha and Zifeng Wang and Tomas Pfister},
      year={2024},
      eprint={2306.00739},
      archivePrefix={arXiv},
      primaryClass={cs.CL},
      url={https://arxiv.org/abs/2306.00739}, 
}

@misc{scholak2021picard,
      title={PICARD: Parsing Incrementally for Constrained Auto-Regressive Decoding from Language Models}, 
      author={Torsten Scholak and Nathan Schucher and Dzmitry Bahdanau},
      year={2021},
      eprint={2109.05093},
      archivePrefix={arXiv},
      primaryClass={cs.CL},
      url={https://arxiv.org/abs/2109.05093}, 
}

@inproceedings{wang2023know,
  title={Know What I don’t Know: Handling Ambiguous and Unknown Questions for Text-to-SQL},
  author={Wang, Bing and Gao, Yan and Li, Zhoujun and Lou, Jian-Guang},
  booktitle={Findings of the Association for Computational Linguistics: ACL 2023},
  pages={5701--5714},
  year={2023}
}

@inproceedings{guo2021chase,
  title={Chase: A large-scale and pragmatic chinese dataset for cross-database context-dependent text-to-sql},
  author={Guo, Jiaqi and Si, Ziliang and Wang, Yu and Liu, Qian and Fan, Ming and Lou, Jian-Guang and Yang, Zijiang and Liu, Ting},
  booktitle={Proceedings of the 59th Annual Meeting of the Association for Computational Linguistics and the 11th International Joint Conference on Natural Language Processing (Volume 1: Long Papers)},
  pages={2316--2331},
  year={2021},
  url={https://github.com/xjtu-intsoft/chase}

}

@inproceedings{liu2022augmenting,
    title = "Augmenting Multi-Turn Text-to-{SQL} Datasets with Self-Play",
    author = "Liu, Qi  and
      Ye, Zihuiwen  and
      Yu, Tao  and
      Song, Linfeng  and
      Blunsom, Phil",
    editor = "Goldberg, Yoav  and
      Kozareva, Zornitsa  and
      Zhang, Yue",
    booktitle = "Findings of the Association for Computational Linguistics: EMNLP 2022",
    month = dec,
    year = "2022",
    address = "Abu Dhabi, United Arab Emirates",
    publisher = "Association for Computational Linguistics",
    url = "https://aclanthology.org/2022.findings-emnlp.411",
    doi = "10.18653/v1/2022.findings-emnlp.411",
    pages = "5608--5620",
}

@misc{wang2024conda,
      title={Controllable Data Augmentation for Context-Dependent Text-to-SQL}, 
      author={Dingzirui Wang and Longxu Dou and Wanxiang Che},
      year={2023},
      eprint={2304.13902},
      archivePrefix={arXiv},
      primaryClass={cs.CL},
      url={https://arxiv.org/abs/2304.13902}, 
}

@article{zhang2024se,
  title={SE-HCL: Schema Enhanced Hybrid Curriculum Learning for Multi-Turn Text-to-SQL},
  author={Zhang, Yiyun and Huang, Gengsheng and others},
  journal={IEEE Access},
  year={2024},
  publisher={IEEE}
}

@article{wei2022chain,
  title={Chain-of-thought prompting elicits reasoning in large language models},
  author={Wei, Jason and Wang, Xuezhi and Schuurmans, Dale and Bosma, Maarten and Xia, Fei and Chi, Ed and Le, Quoc V and Zhou, Denny and others},
  journal={Advances in neural information processing systems},
  volume={35},
  pages={24824--24837},
  year={2022}
}

@inproceedings{bertomeu2006contextual,
  title={Contextual phenomena and thematic relations in database QA dialogues: results from a Wizard-of-Oz experiment},
  author={Bertomeu, N{\'u}ria and Uszkoreit, Hans and Frank, Anette and Krieger, Hans-Ulrich and J{\"o}rg, Brigitte},
  booktitle={Proceedings of the Interactive Question Answering Workshop at HLT-NAACL 2006},
  pages={1--8},
  year={2006}
}

@misc{zhang2020did,
      title={Did You Ask a Good Question? A Cross-Domain Question Intention Classification Benchmark for Text-to-SQL}, 
      author={Yusen Zhang and Xiangyu Dong and Shuaichen Chang and Tao Yu and Peng Shi and Rui Zhang},
      year={2020},
      eprint={2010.12634},
      archivePrefix={arXiv},
      primaryClass={cs.CL},
      url={https://arxiv.org/abs/2010.12634}, 
}

@inproceedings{yu2019sparc,
    title = "{SP}ar{C}: Cross-Domain Semantic Parsing in Context",
    author = "Yu, Tao  and
      Zhang, Rui  and
      Yasunaga, Michihiro  and
      Tan, Yi Chern  and
      Lin, Xi Victoria  and
      Li, Suyi  and
      Er, Heyang  and
      Li, Irene  and
      Pang, Bo  and
      Chen, Tao  and
      Ji, Emily  and
      Dixit, Shreya  and
      Proctor, David  and
      Shim, Sungrok  and
      Kraft, Jonathan  and
      Zhang, Vincent  and
      Xiong, Caiming  and
      Socher, Richard  and
      Radev, Dragomir",
    editor = "Korhonen, Anna  and
      Traum, David  and
      M{\`a}rquez, Llu{\'\i}s",
    booktitle = "Proceedings of the 57th Annual Meeting of the Association for Computational Linguistics",
    month = jul,
    year = "2019",
    address = "Florence, Italy",
    publisher = "Association for Computational Linguistics",
    url = "https://aclanthology.org/P19-1443",
    doi = "10.18653/v1/P19-1443",
    pages = "4511--4523",
}

@article{madaan2024self,
  title={Self-refine: Iterative refinement with self-feedback},
  author={Madaan, Aman and Tandon, Niket and Gupta, Prakhar and Hallinan, Skyler and Gao, Luyu and Wiegreffe, Sarah and Alon, Uri and Dziri, Nouha and Prabhumoye, Shrimai and Yang, Yiming and others},
  journal={Advances in Neural Information Processing Systems},
  volume={36},
  year={2024}
}

@misc{wu2024stateflow,
      title={StateFlow: Enhancing LLM Task-Solving through State-Driven Workflows}, 
      author={Yiran Wu and Tianwei Yue and Shaokun Zhang and Chi Wang and Qingyun Wu},
      year={2024},
      eprint={2403.11322},
      archivePrefix={arXiv},
      primaryClass={cs.CL},
      url={https://arxiv.org/abs/2403.11322}, 
}

@article{dettmers2024qlora,
  title={Qlora: Efficient finetuning of quantized llms},
  author={Dettmers, Tim and Pagnoni, Artidoro and Holtzman, Ari and Zettlemoyer, Luke},
  journal={Advances in Neural Information Processing Systems},
  volume={36},
  year={2024}
}

@article{li2024can,
  title={Can llm already serve as a database interface? a big bench for large-scale database grounded text-to-sqls},
  author={Li, Jinyang and Hui, Binyuan and Qu, Ge and Yang, Jiaxi and Li, Binhua and Li, Bowen and Wang, Bailin and Qin, Bowen and Geng, Ruiying and Huo, Nan and others},
  journal={Advances in Neural Information Processing Systems},
  volume={36},
  year={2024}
}

@inproceedings{hui2021dynamic,
  title={Dynamic hybrid relation exploration network for cross-domain context-dependent semantic parsing},
  author={Hui, Binyuan and Geng, Ruiying and Ren, Qiyu and Li, Binhua and Li, Yongbin and Sun, Jian and Huang, Fei and Si, Luo and Zhu, Pengfei and Zhu, Xiaodan},
  booktitle={Proceedings of the AAAI conference on artificial intelligence},
  volume={35},
  pages={13116--13124},
  year={2021}
}

@misc{lee2024trustsql,
      title={TrustSQL: Benchmarking Text-to-SQL Reliability with Penalty-Based Scoring}, 
      author={Gyubok Lee and Woosog Chay and Seonhee Cho and Edward Choi},
      year={2024},
      eprint={2403.15879},
      archivePrefix={arXiv},
      primaryClass={cs.AI},
      url={https://arxiv.org/abs/2403.15879}, 
}

@inproceedings{guo2018question,
    title = "Question Generation from {SQL} Queries Improves Neural Semantic Parsing",
    author = "Guo, Daya  and
      Sun, Yibo  and
      Tang, Duyu  and
      Duan, Nan  and
      Yin, Jian  and
      Chi, Hong  and
      Cao, James  and
      Chen, Peng  and
      Zhou, Ming",
    editor = "Riloff, Ellen  and
      Chiang, David  and
      Hockenmaier, Julia  and
      Tsujii, Jun{'}ichi",
    booktitle = "Proceedings of the 2018 Conference on Empirical Methods in Natural Language Processing",
    month = oct # "-" # nov,
    year = "2018",
    address = "Brussels, Belgium",
    publisher = "Association for Computational Linguistics",
    url = "https://aclanthology.org/D18-1188",
    doi = "10.18653/v1/D18-1188",
    pages = "1597--1607",
}

@software{numbersstation2023NSText2SQL,
  author    = {NumbersStationAI},
  title     = {NSText2SQL: An Open Source Text-to-SQL Dataset for Foundation Model Training},
  month     = {July},
  year      = {2023},
  url       = {https://github.com/NumbersStationAI/NSQL},
}

@software{defog2023sql,
  author    = {DefogAI},
  title     = {Defog SQLCoder},
  year      = {2023},
  url       = {https://github.com/defog-ai/sqlcoder},
}

@misc{xu2023wizardlmempoweringlargelanguage,
      title={WizardLM: Empowering Large Language Models to Follow Complex Instructions}, 
      author={Can Xu and Qingfeng Sun and Kai Zheng and Xiubo Geng and Pu Zhao and Jiazhan Feng and Chongyang Tao and Daxin Jiang},
      year={2023},
      eprint={2304.12244},
      archivePrefix={arXiv},
      primaryClass={cs.CL},
      url={https://arxiv.org/abs/2304.12244}, 
}

@inproceedings{NEURIPS2023_91f18a12,
 author = {Zheng, Lianmin and Chiang, Wei-Lin and Sheng, Ying and Zhuang, Siyuan and Wu, Zhanghao and Zhuang, Yonghao and Lin, Zi and Li, Zhuohan and Li, Dacheng and Xing, Eric and Zhang, Hao and Gonzalez, Joseph E and Stoica, Ion},
 booktitle = {Advances in Neural Information Processing Systems},
 editor = {A. Oh and T. Naumann and A. Globerson and K. Saenko and M. Hardt and S. Levine},
 pages = {46595--46623},
 publisher = {Curran Associates, Inc.},
 title = {Judging LLM-as-a-Judge with MT-Bench and Chatbot Arena},
 url = {https://proceedings.neurips.cc/paper_files/paper/2023/file/91f18a1287b398d378ef22505bf41832-Paper-Datasets_and_Benchmarks.pdf},
 volume = {36},
 year = {2023}
}

@inproceedings{ding-etal-2024-data,
    title = "Data Augmentation using {LLM}s: Data Perspectives, Learning Paradigms and Challenges",
    author = "Ding, Bosheng  and
      Qin, Chengwei  and
      Zhao, Ruochen  and
      Luo, Tianze  and
      Li, Xinze  and
      Chen, Guizhen  and
      Xia, Wenhan  and
      Hu, Junjie  and
      Luu, Anh Tuan  and
      Joty, Shafiq",
    editor = "Ku, Lun-Wei  and
      Martins, Andre  and
      Srikumar, Vivek",
    booktitle = "Findings of the Association for Computational Linguistics ACL 2024",
    month = aug,
    year = "2024",
    address = "Bangkok, Thailand and virtual meeting",
    publisher = "Association for Computational Linguistics",
    url = "https://aclanthology.org/2024.findings-acl.97",
    doi = "10.18653/v1/2024.findings-acl.97",
    pages = "1679--1705",
}

@misc{yu2021grappagrammaraugmentedpretrainingtable,
      title={GraPPa: Grammar-Augmented Pre-Training for Table Semantic Parsing}, 
      author={Tao Yu and Chien-Sheng Wu and Xi Victoria Lin and Bailin Wang and Yi Chern Tan and Xinyi Yang and Dragomir Radev and Richard Socher and Caiming Xiong},
      year={2021},
      eprint={2009.13845},
      archivePrefix={arXiv},
      primaryClass={cs.CL},
      url={https://arxiv.org/abs/2009.13845}, 
}

@misc{saparina2024ambrosiabenchmarkparsingambiguous,
      title={AMBROSIA: A Benchmark for Parsing Ambiguous Questions into Database Queries}, 
      author={Irina Saparina and Mirella Lapata},
      year={2024},
      eprint={2406.19073},
      archivePrefix={arXiv},
      primaryClass={cs.CL},
      url={https://arxiv.org/abs/2406.19073}, 
}

@INPROCEEDINGS{10598154,
  author={Ganti, Manasi and Orr, Laurel and Wu, Sen},
  booktitle={2024 IEEE 40th International Conference on Data Engineering (ICDE)}, 
  title={Evaluating Text-to-SQL Model Failures on Real-World Data}, 
  year={2024},
  volume={},
  number={},
  pages={1-1},
  doi={10.1109/ICDE60146.2024.00456}}

@misc{saparina2024,
      title={AMBROSIA: A Benchmark for Parsing Ambiguous Questions into Database Queries}, 
      author={Irina Saparina and Mirella Lapata},
      year={2024},
      eprint={2406.19073},
      archivePrefix={arXiv},
      primaryClass={cs.CL},
      url={https://arxiv.org/abs/2406.19073}, 
}

@misc{iourovitski2024gradescorequantifyingllm,
      title={Grade Score: Quantifying LLM Performance in Option Selection}, 
      author={Dmitri Iourovitski},
      year={2024},
      eprint={2406.12043},
      archivePrefix={arXiv},
      primaryClass={cs.AI},
      url={https://arxiv.org/abs/2406.12043}, 
}

\end{document}